%% file: main.tex
\documentclass[10pt,twocolumn,letterpaper]{article}

\usepackage[pagenumbers]{cvpr} % To force page numbers, e.g. for an arXiv version

\usepackage{multirow}
\usepackage{enumitem}
\usepackage[table]{xcolor} % 用于表格颜色
\usepackage{tabularx}     % 用于 \begin{tabularx}
\usepackage{array}

\definecolor{PastelGreen}{HTML}{D0E8D0} % 原 F0F9F0
\newcolumntype{C}{>{\centering\arraybackslash}X}
\definecolor{PastelYellow}{HTML}{F5F5D8} % 原 FCFCEB
\input{preamble}
\definecolor{cvprblue}{rgb}{0.21,0.49,0.74}
\usepackage[pagebackref,breaklinks,colorlinks,allcolors=cvprblue]{hyperref}

\def\paperID{****} % *** Enter the Paper ID here
\def\confName{****}
\def\confYear{****}

\title{Open-world Hand-Object Interaction Video Generation Based on Structure and Contact-aware Representation}

\author{
Haodong Yan$^{1}$ \quad
Hang Yu$^{1}$ \quad
Zhide Zhong$^{1}$ \quad
Weilin Yuan$^{1}$ \quad
Xin Gong$^{1}$ \quad
Zehang Luo$^{1}$ \\
Chengxi Heyu$^{1}$ \quad
Junfeng Li$^{1}$ \quad
Wenxuan Song$^{1}$ \quad
Shunbo Zhou$^{2}$ \quad
Haoang Li$^{1}$ \\[0.5em] % 在名字和单位之间稍微增加一点空隙
{% 字体变小，如果觉得不够小可以用 \footnotesize
$^{1}$The Hong Kong University of Science and Technology (Guangzhou) \quad % 这里用 \quad 分隔
$^{2}$Huawei Cloud
}
}

\begin{document}
\twocolumn[{%
\renewcommand\twocolumn[1][]{#1}%
\maketitle
\centering
\includegraphics[width=.98\linewidth]{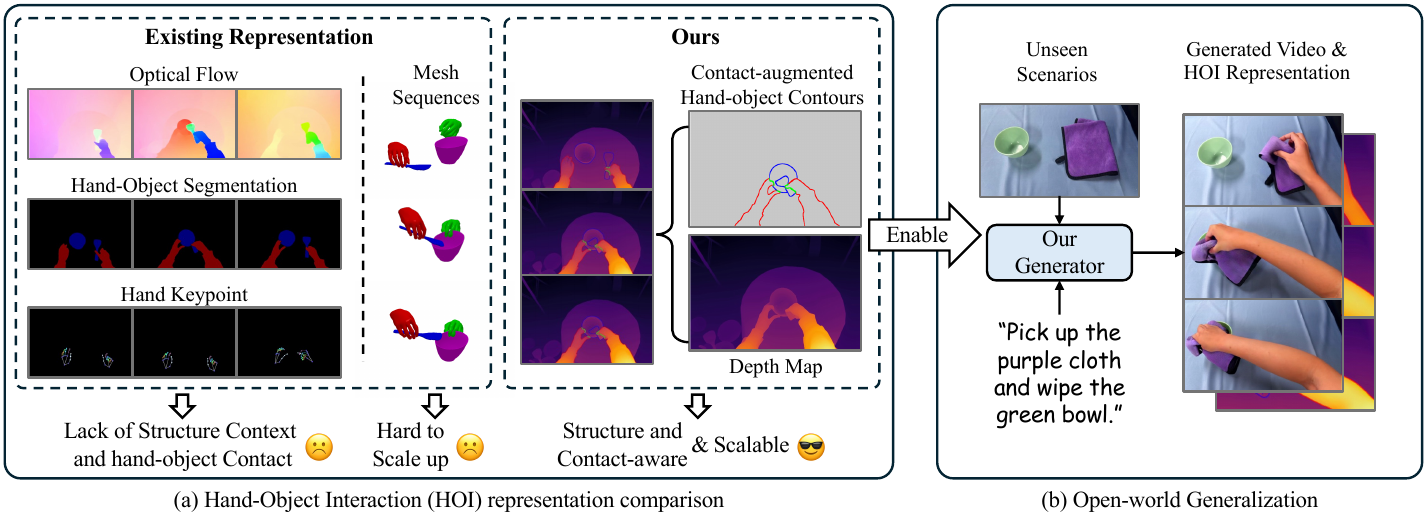}
\captionof{figure}{Overview of our structure and
contact-aware representation and its enabled open-world generalization. (a) Prior HOI representations lack either scalability (e.g., 3D mesh) or crucial contact/structure cues (e.g., optical flow or segmentation). Our approach resolves this dilemma with a representation composed of two scalable and complementary components: 1) contact-augmented hand-object contours for capturing the contact region and hand-object spatial localization, and 2) depth maps offering holistic structure context. (b) Our structure and contact-aware representation acts as an additional interaction-oriented generative supervision signal. By learning to jointly generate videos and our representation at a large scale, our model captures interaction patterns consistent with physical constraints, enabling strong generalization to complex open-world interactions, even with unseen non-rigid objects. \vspace{1em}}
\label{fig:teaser}
}]
\input{sec/0_abstract}    
\input{sec/1_intro}

\input{sec/2_related_work}
\input{sec/3_method}

\input{sec/4_experiment}
\input{sec/5_conclusion}
{
    \small
    \bibliographystyle{ieeenat_fullname}
    \bibliography{main}
}

% WARNING: do not forget to delete the supplementary pages from your submission 
%\input{sec/X_suppl}

% WARNING: do not forget to delete the supplementary pages from your submission 
%\input{sec/X_suppl}

\end{document}

%% file: sec/0_abstract.tex
\begin{abstract}
Generating realistic hand-object interactions (HOI) videos is a significant challenge due to the difficulty of modeling physical constraints (e.g., contact and occlusion between hands and manipulated objects). Current methods utilize HOI representation as an auxiliary generative objective to guide video synthesis. However, there is a dilemma between 2D and 3D representations that cannot simultaneously guarantee scalability and interaction fidelity. To address this limitation, we propose a structure and contact-aware representation that captures hand-object contact, hand-object occlusion, and holistic structure context without 3D annotations. This interaction-oriented and scalable supervision signal enables the model to learn fine-grained interaction physics and generalize to open-world scenarios. To fully exploit the proposed representation, we introduce a joint-generation paradigm with a share-and-specialization strategy that generates interaction-oriented representations and videos. Extensive experiments demonstrate that our method outperforms state-of-the-art methods on two real-world datasets in generating physics-realistic and temporally coherent HOI videos. Furthermore, our approach exhibits strong generalization to challenging open-world scenarios, highlighting the benefit of our scalable design. Our project page is \url{https://hgzn258.github.io/SCAR/}.
\end{abstract}

%% file: sec/1_intro.tex
\section{Introduction}
%\begin{figure}[t]
%    \centering

    % --- First Subfigure ---
    %\begin{subfigure}{\linewidth}
    %    \centering
%        \includegraphics[width=\linewidth]{images/teaser_a.pdf} % Replace with your image path
    %     \caption{Hand-Object Interaction (HOI) representations comparison.}
        %\label{fig:teaser_sub1}
    %\end{subfigure}
    %\vspace{1ex} % Adds a small vertical space between the subfigures

    % --- Second Subfigure ---
    %\begin{subfigure}{\linewidth}
    %    \centering
     %   \includegraphics[width=\linewidth]{images/teaser_b.pdf} % Replace with your image path
     %   \caption{Generation paradigms comparison.}
      %  \label{fig:teaser_sub2}
    %\end{subfigure}

    % --- Main Figure Caption ---
%    \caption{Hand-Object Interaction (HOI) representations comparison. Existing HOI representations lack either contact and structure cues (left) or scalability (middle). Our approach (right) resolves this dilemma with a representation composed of two scalable 2D components: 1) a contact-augmented hand-object contours for capturing the contact region and hand-object spatial localization, and 2) a depth map offering holistic structure context.} %(b) Prior multi-stage pipelines (left) suffer from error accumulation. In contrast, our end-to-end joint-generation paradigm (right) alleviates this by simultaneously generating the video and HOI representation.}
%    \label{fig:teaser}
%\end{figure}
%\haodong{HOI-HOI video generation - General I2V  - Introducing hoi representation - 3d hoi representation - multi-stag - ours} 
%The dexterous manipulation of objects is a key aspect of human intelligence. 
Achieving a deep understanding and enabling realistic synthesis of this complex hand-object interaction (HOI) is a key challenge in computer vision. This domain has wide-ranging applications in robotic learning~\cite{zhao2025taste, hu2024}, augmented reality~\cite{hu2022hand, pei2022hand}, and human behavior analysis~\cite{gupta2009observing,fan2021understanding}. A particularly challenging frontier for this task is HOI video generation. Conditioned on an observed image and a task description, the goal is to synthesize a video sequence of hands manipulating objects with physics-realistic interaction and temporally coherent motion. Although recent video generation models~\cite{hongcogvideo,shi2024motion,wan2025wan,yang2024cogvideox,renconsisti2v,wang2025transpixeler} have made progress in photorealism and visual fidelity, their success does not readily transfer to the nuanced domain of HOI generation. The lack of specific inductive biases (e.g., hand-object contact and holistic structure) causes these general-purpose models to often fail on physics realism. 
%These general-purpose models lack the specific inductive biases that often fail on physics realism (e.g., hand-object contact and holistic structure). %These general-purpose models lack the specific inductive biases to model crucial physical interaction constraints (e.g., hand-object contact, holistic structure and hand-object motion). As a result, they often produce physically implausible outcomes. 

One line of work~\cite {zhao2025taste, li2025mask2iv} attempts to tackle this problem by leveraging an HOI representation that explicitly models physical interaction cues as an auxiliary objective. Despite this advance, the design of the HOI representation presents a dilemma between scalability and interaction fidelity (see~\cref{fig:teaser}(a)). Scalable 2D representations such as optical flow~\cite{shi2024motion, jin2025flovd}, hand-object segmentation~\cite{yariv2025through,li2025mask2iv}, and hand keypoint~\cite{zhao2025taste} lack a holistic structure context and a hand-object contact region. Conversely, 3D mesh sequences~\cite{cha2024text2hoi,huang2025hoigpt, pang2025manivideo,dang2025svimo} with complete structure context suffer from poor scalability due to costly 3D annotations. In addition, these approaches~\cite{shi2024motion, jin2025flovd, zhao2025taste, yariv2025through, li2025mask2iv} follow a multi-stage paradigm, where each stage is trained with ground-truth inputs yet conditioned on predictions from the previous stage at inference, causing accumulated errors that degrade both physics realism and visual quality~\cite{kang2022error,jin2025flovd,wang2025error}.

To address the aforementioned scalability-fidelity dilemma, we introduce a novel representation whose overall design is depicted in~\cref{fig:teaser}(a). It consists of two complementary and scalable components: 1) a contact-augmented hand-object contour that explicitly captures hand-object contact and spatial localization, and 2) a depth map that provides the holistic structure context. This structure and contact-aware representation thus serves as a scalable and interaction-oriented supervisory signal. It guides the model to learn interaction patterns consistent with physical constraints, thereby promoting both physics-realistic generation and open-world generalization (see~\cref{fig:teaser}(b)).  %This pipeline (detailed in~\cref{subsec:hoi_representation}) begins with hand-object segmentation extraction, where CoT-guided vision-language models perform robust grounding. Next, a powerful proxy for hand-object contact is derived from the intersection of dilated hand-object mask contours. The depth map is obtained from a consistent video depth estimator~\cite{chen2025video}. Finally, these components are alpha-blended to form the final representation. %It uses CoT-guided vision-language models for robust hand-object grounding and a consistent video depth estimator~\cite{chen2025video} to obtain depth maps.

%To support large-scale learning, we develop a data curation pipeline to generate proposed representations over 100k HOI videos without costly 3D annotations. 
Briefly, we generate the above representation as follows. To create the contact-augmented hand-object contour, we first use a vision-language model (VLM)~\cite{bai2025qwen2} with a sophisticated Chain-of-Thought (CoT)~\cite{wei2022chain} prompt to ground hands and objects. Subsequently, we leverage SAM2~\cite{ravi2024sam} to extract and propagate hand-object masks, from which we derive an elegant and powerful proxy for the hand-object contact region by computing the intersection of their dilated contours. For the second component, the depth map is obtained from a video-consistent depth estimator~\cite{chen2025video}. %To support large-scale learning, we develop a scalable data curation pipeline to generate proposed representations over 100k HOI videos without costly 3D supervision. 
%By reasoning over the task description and video frames, the model accurately disambiguates and grounds the manipulated object. Subsequently, we leverage SAM2~\cite{ravi2024sam} to extract and propagate hand-object masks, which provide explicit spatial localization. Our key insight is that a powerful proxy for the hand-object contact region can be derived from these masks by computing the intersection of their dilated contours. For the second component, the depth map is obtained from a video depth estimator~\cite{chen2025video}. 
%These components are fused by alpha-blending the contour onto the depth map, yielding a structure and contact-aware representation. 

Based on the above representation, we propose a joint-generation paradigm as shown in~\cref{fig:joint_generation}. Different from mainstream multi-stage approaches~\cite{shi2024motion, jin2025flovd, zhao2025taste, yariv2025through, li2025mask2iv}, our method generates proposed HOI representations and videos simultaneously to mitigate error accumulation. This paradigm is enabled by a hierarchical joint denoiser, which learns the joint distribution between videos and their structure and contact-aware representations through two cascaded modules: 1) a \textit{Shared Semantics} module that captures shared, modality-invariant semantics by enforcing alignment via an explicit regularization loss on hidden states, and 2) a \textit{Specialized Details} module that removes this constraint to capture unique, modality-specific characteristics. By integrating the above techniques, our method achieves physics-realistic HOI video generation.  %Through large-scale training that jointly generates videos along with proposed representations, our model captures fine-grained interaction physics, thereby achieving physics-realistic HOI video generation with strong generalization to open-world scenarios. 
To summarize, our contributions are as follows:
\begin{itemize}
    \item We propose a structure and contact-aware representation as a scalable and interaction-oriented supervisory signal that guides the model to capture fine-grained interaction physics. We curate this representation for over 100k HOI videos, facilitating large-scale training. 
    \item We introduce a joint-generation paradigm with a share-and-specialization strategy that generates proposed HOI representations and videos simultaneously, mitigating multi-stage error accumulation.
    \item Extensive experiments demonstrate our method generates physics-realistic HOI videos, surpassing state-of-the-art methods on two real-world datasets and showing strong generalization to open-world scenarios. 
\end{itemize}

%: 1) a more informative auxiliary interaction cue that integrates hand-object occlusion relationships, scene depths, and contact regions; and 2) a unified generation paradigm in which video frames and interaction cues are predicted jointly, reducing error accumulation and enhancing physical interaction fidelity.

%% file: sec/2_related_work.tex
\section{Related Work}

\begin{figure*}[!t]
  \centering
    \includegraphics[width=0.99\linewidth]{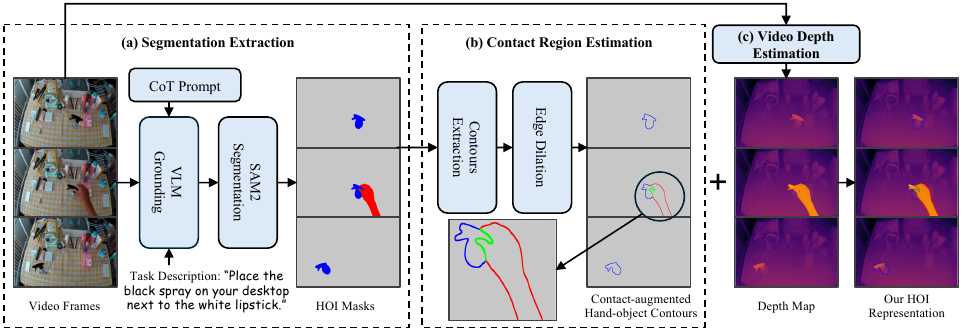}
    \caption{%Our structure and contact-aware HOI representation is constructed by alpha-blending contact-augmented hand-object contours onto video depth maps. %, creating a representation that is simultaneously rich in hand-object contact while preserving holistic structure context. 
    Overview of our structure and contact-aware representation curation pipeline. It begins with (a) \textbf{Segmentation Extraction}, where a CoT-guided VLM grounds hand and object from the input RGB video, and SAM2 generates HOI masks. Next, (b) \textbf{Contact Region Estimation} produces the final contact-augmented hand-object contours by computing a contact region from the intersection of the dilated hand and object contours. In parallel, (c) \textbf{Video Depth Estimation} generates a dense depth map sequence for holistic structure. Finally, these contact-augmented hand-object contours are alpha-blended onto the depth maps to form the final HOI representation.}
    \label{fig:hoi_curation}
\end{figure*}
\subsection{Hand-Object Interaction Representation}
\label{sec:hoi_rep}
Hand-object interaction (HOI) representation aims to capture both the individual dynamics of hands and objects and their physical correlations (e.g., contact and occlusion). In HOI generation and prediction tasks~\cite{cha2024text2hoi,huang2025hoigpt,zhou2025megohand,luo2025being,dang2025svimo}, interactions are commonly represented as sequences of MANO parameters~\cite{MANO:SIGGRAPHASIA:2017} or 3D hand keypoints, along with object meshes or point clouds. While offering fine-grained structure detail for physical realism, these high-fidelity representations lack explicit signals for crucial interaction semantics, such as physical contact. To bridge this gap, some works further extract interaction-centric features directly from this 3D data. HOI-GEN~\cite{hu2022hand} and ManiVideo~\cite{pang2025manivideo} introduce occlusion-aware topology maps as HOI representation, which improves depth and occlusion consistency and enriches spatial control. Another line of work~\cite{grady2021contactopt,jiang2021hand,karunratanakul2020grasping,zhou2022toch,liu2023contactgen} utilizes the contact map as the primary HOI representation. These methods explicitly model regions of physical interaction between the hand and object, offering a focused representation for contact semantics. %such as contact maps~\cite{liu2023contactgen} and topology maps for occlusion~\cite{hu2022hand, pang2025manivideo}. 
%However, because all these high-fidelity representations are derived from or supervised by the underlying 3D structure, they inherit a critical scalability bottleneck, relying on expensive 3D supervision (e.g., motion capture) that is infeasible for large, in-the-wild datasets. 
However, these representations with high interaction fidelity rely on expensive 3D annotations (e.g., motion capture) that are hard to scale up. 
To address this issue, an alternative approach leverages scalable motion representations like overlapped masks~\cite{yariv2025through,li2025mask2iv}, optical flow~\cite{shi2024motion,jin2025flovd}, or 2D hand keypoints~\cite{zhao2025taste,bao2024handsonvlm}. %. To capture manipulation-centric semantics, a series of works~\cite{zhao2025taste,bao2024handsonvlm} also specifically leverage 2D hand keypoints as a representation. %\haodong{mention interaction affordances? (not complete)} 
%Some of them~\cite{bahl2023affordances,liu2022joint} further propose interaction affordances based on fingertip keypoints on the object within a single interaction frame. 
%However, by confining contact to a sparse set of fingertips, these methods neglect the crucial role of other hand regions, such as the palm. 
Despite their scalability, these representations fail to capture hand-object contact regions and a holistic structure context in the scene. In contrast, we introduce a structure and contact-aware representation that encodes hand-object contact, hand-object spatial localization, and holistic structure context without 3D annotations.  %These limitations motivate our approach of proposing a more informative HOI representation that completely encodes a hand-object contact region, hand-object spatial localization, and holistic structure context derived directly from 2D video to maintain scalability.
\subsection{Hand-Object Interaction Video Generation}
%The synthesis of dynamic human-object interactions is a nascent and critical research area with significant implications for robotics, AR/VR, and human behavior analysis. This domain directly benefits from recent advances in general video generation~\cite{yang2024cogvideox,wan2025wan,wang2025transpixeler,kong2024hunyuanvideo}, driven primarily by powerful Diffusion Transformer (DiT) architectures~\cite{peebles2023scalable} that model spatio-temporal correlations in a compressed latent space. %In particular, the ability to generate dynamic HOI videos offers a detailed and intuitive demonstration.
The synthesis of the human-object interaction (HOI) has recently benefited from significant advances in video generation~\cite{yang2024cogvideox,wan2025wan,wang2025transpixeler,kong2024hunyuanvideo} with powerful Diffusion Transformer (DiT) architectures~\cite{peebles2023scalable}. Current HOI video generation research can be broadly categorized into representation-conditioned and task-guided settings. The former methods~\cite{pang2025manivideo,xu2024anchorcrafter, wang2025dreamactor,fan2025re} generate HOI videos by conditioning on pre-existing HOI representations. %However, these methods face significant input constraints, as these representations must be explicitly pre-defined or extracted from a driving video. 
However, these methods face significant input constraints, as their reliance on control signals must be manually predefined or extracted from a driving video.
In contrast, the task-guided methods~\cite{li2025mask2iv,gan2025humandit,zhao2025taste} generate HOI videos conditioned on textual commands, which are more readily available and practical for real-world applications. For example, VPP~\cite{hu2024} directly fine-tunes SVD~\cite{blattmann2023stable} on HOI data to adapt the domain. MaskI2V~\cite{li2025mask2iv} first predicts an interaction trajectory represented by a sequence of masks, and then uses this trajectory to guide the final video generation. Taste-Rob~\cite{zhao2025taste} first generates a coarse video to extract an initial hand pose sequence. This sequence is then refined by a motion diffusion model and used as a condition for the final video generation.

However, existing methods struggle with physics realism and the generalization ability to open-world scenarios, primarily due to the representation dilemma (\cref{sec:hoi_rep}). In addition, these approaches often rely on a multi-stage paradigm, which is susceptible to error accumulation. In contrast, our work employs a joint-generation paradigm to mitigate error accumulation and capture the inherent correlation between HOI representation and video.

%% file: sec/3_method.tex
\section{Method}
\begin{figure*}[!t]
    \centering
    \includegraphics[width=0.98\linewidth]{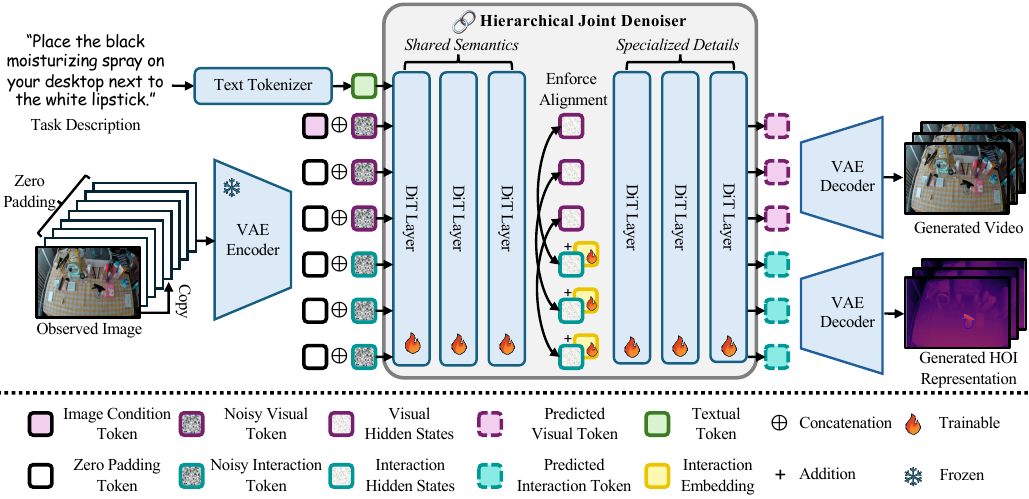}
    \caption{The joint-generation paradigm of our method. %Ground-truth RGB videos and their HOI representations are encoded into clean latent tokens by a VAE. These tokens are concatenated, then corrupted by noise in the forward diffusion process. 
    %Our framework takes an observed image and a text prompt as primary inputs to generate the HOI representation and video jointly. The text prompt is tokenized into textual tokens. The observed image is first padded with future blank frames and passed through a frozen VAE encoder to generate condition tokens. They are then concatenated with their respective noisy visual and interaction tokens to serve as a visual condition. 
    Given an observed image and a task description, our framework jointly generates a video and its corresponding HOI representation. The core technical novelty lies in the \textbf{Hierarchical Joint Denoiser} that co-denoises visual and interaction tokens within a unified latent space. First, the \textit{Shared Semantics} module enforces cross-modal consistency via an alignment loss (maximizing cosine similarity) to capture shared semantics like spatial layout and temporal dynamics. Then, the \textit{Specialized Details} module adds a learnable interaction embedding to capture modality-specific details. Finally, the denoised predicted visual and interaction tokens are passed through the VAE decoder to reconstruct both outputs. }
\label{fig:joint_generation}
\end{figure*}
%The overall pipeline of our method is illustrated in \cref{fig:onecol}.Given a reference image \(x^0 \in \mathbb{R}^{H \times W \times 3}\) and a HOI task description \(T\), our method generates video frames
%together with synchronized interaction cues $\{(x^n,I^n)\}_i^N$, where \(x^n, I^n \in \mathbb{R}^{H \times W \times 3}\). 
%In this section, we first review the preliminaries of DIT-based video generation~\cite{yang2024cogvideox,wan2025wan} in \cref{subsec:preliminary}. Next, we define our novel HOI representation and present its curation pipeline in \cref{subsec:hoi_representation}. Finally, we describe our joint-generation paradigm in \cref{subsec:joint_generation_model}.
In this section, we first introduce our structure and contact-aware HOI representation in \cref{subsec:hoi_representation}. Then, we detail our joint-generation paradigm in \cref{subsec:joint_generation_model} to generate the proposed HOI representation and videos simultaneously.

 %These conditions are first passed through their respective encoders (e.g., a text encoder) to produce a sequence of conditioning embeddings $E_c$. These embeddings are then fused into the DiT, typically via cross-attention. The network itself is a stack of $K$ transformer blocks, where the update rule for the hidden states of the $k$-th layer is:
%$$H^{(k)} = \mathcal{B}^{(k)}\bigl(H^{(k-1)}, E_c, t\bigr), \quad k = 1, \dots, K.$$ 
%Each block combines spatio-temporal self-attention for modeling latent dynamics and cross-attention for incorporating prompts. 
%CogVideoX~\cite{yang2024cogvideox} employs expert AdaLN layers and full 3D self-attention to fuse text and video latents, while Wan~2.1~\cite{wan2025wan} emphasizes causal compression and integrates explicit cross-attention to prompt embeddings, highlighting different trade-offs in DiT-based frameworks.

\subsection{Structure and Contact-aware Representation}
\label{subsec:hoi_representation}
As illustrated in~\cref{fig:teaser}(a), our representation is formed by alpha-blending two scalable and complementary components: (1) contact-augmented contours that encode hand-object contact and hand-object spatial localization, and (2) depth maps providing holistic structure context. We chose sparse contours instead of dense masks to capture hand-object spatial localization because sparse contours preserve the depth map during alpha-blending, while dense masks tend to obscure this critical information. The full curation pipeline is detailed below.

\noindent \textbf{Segmentation Extraction.} As depicted in~\cref{fig:hoi_curation}(a), we adopt a grounding–and-tracking pipeline to obtain hand and object masks. Specifically, we first localize the hand and object using a large VLM~\cite{bai2025qwen2} prompted with Chain-of-Thought (CoT) reasoning. The CoT prompt guides the VLM to sequentially verify textual intent, visual interaction cues, and temporal motion, enabling more reliable grounding than specialized detectors~\cite{liu2024grounding}, especially for open-vocabulary objects and complex scenes involving distractors. Details of VLM-based grounding are provided in the supplementary material. Then, we use the grounded bounding boxes to prompt SAM2~\cite{ravi2024sam}, which extracts and propagates them into per-frame hand and object masks. To guarantee the final annotation quality, this automated process was followed by a manual verification stage, where human annotators review and correct the generated masks.

\noindent\textbf{Contact Region Estimation.} 
Given the hand and object masks, we estimate the hand-object contact region by identifying areas where their boundaries exhibit close spatial proximity (see~\cref{fig:hoi_curation}(b)). Specifically, we first convert these masks into thin contours, denoted as $E_\textnormal{h}$ and $E_\textnormal{o}$. We then estimate hand-object contact regions by dilating the hand with a fixed radius, $r_\textnormal{h}$, and the object with a scale-adaptive radius, $r_\textnormal{o}$, to robustly handle its significant scale variations. To ensure robustness to extreme object scales, $r_\textnormal{o}$ is calculated proportionally to the diagonal length of its bounding box $L$ and constrained in a reasonable range $[r_{\min}, r_{\max}]$:
\begin{equation}
r_\textnormal{o} = \min(r_{\max}, \max(r_{\min}, \beta \cdot L )).
\end{equation}
Here, $\beta$ is a proportional coefficient. The hand-object contact region, $C$, is defined as the intersection of these dilated contours:
\begin{equation}
C =\operatorname{dilate}(E_\textnormal{hand}, r_\textnormal{h})\;\cap\;\operatorname{dilate}(E_\textnormal{object}, r_\textnormal{o}).
\end{equation}
This simple yet highly scalable proxy provides reliable estimates of contact regions. The effectiveness of this estimated contact region is demonstrated in~\cref{sec:ana_model}. %Our ablation study (Sec. 4.4, Tab. 2) validates its effectiveness for learning interaction physics.
%\noindent \textbf{Contact-Region Estimation.} 
%Instead of relying on interior overlap or 3D supervision, we operate on \emph{boundaries}: true contact manifests where the hand and object boundaries lie within a small pixel distance. We first convert each mask into a thin contour (morphological edge). We realize this by dilating the two edges and taking their intersection:
%\begin{equation}
%C \;=\; \operatorname{dilate}(E_\textnormal{hand}, r)\;\cap\;\operatorname{dilate}(E_\textnormal{object}, r),
%\end{equation}
%where $E_\textnormal{hand}$ and $E_\textnormal{object}$ is the hand edge and object edge, respectively. $r$ represents the dilation kernel size.
%\noindent \textbf{Hand Mask Extraction.} To generate hand mask annotations, we first select the middle frame of each video and process it with VISOR-HOS, which detects hand regions. The detected hand masks are then used to initialize the Segment Anything Model v2 (SAM2)\cite{ravi2024sam}, which propagates the hand region across all frames, resulting in a dense per-frame mask sequence for the hand.  

\noindent \textbf{Video Depth Estimation.}
To complement our HOI representation with 3D structure context, we incorporate per-frame depth maps (see~\cref{fig:hoi_curation}(c)). We generate these using a state-of-the-art video-consistent depth estimation model~\cite{chen2025video}, which provides crucial information on relative depth ordering. While such models are often scale-ambiguous, their estimated relative structure is highly reliable. Our curated representation leverages this robust signal, providing a scalable 3D structure context independent of absolute scale. %We employ a state-of-the-art video depth estimation model~\cite{chen2025video} to process the input RGB video sequence and generate a corresponding sequence of dense depth maps. %This approach avoids the need for specialized hardware (e.g., RGB-D sensors) and allows our pipeline to be applied to a wide variety of open-world videos, ensuring scalability and broad applicability. The resulting depth stream serves as the foundational layer of our final HOI representation.

\subsection{Joint Generation of Video and Representation}
\label{subsec:joint_generation_model}
%To leverage our constructed HOI representation as interaction cues to better guide HOI video generation, we propose a joint-generation paradigm that minimally adapts the backbone of standard video diffusion models. Specifically, rather than treating interaction cues as a post-processing condition, we inject interaction tokens directly into the diffusion network, enabling the model to reason about RGB motion and interaction simultaneously. Importantly, our modification is plug-and-play and general, allowing for easy integration into various existing video diffusion models.
To fully exploit our representation without multi-stage error accumulation, we propose a novel joint-generation paradigm (see~\cref{fig:joint_generation}). %Our key insight is that the HOI representation acts as an interaction-oriented supervisory signal, providing a ``shortcut’’ for learning interaction physics.  
By compelling the model to simultaneously generate both HOI representation and the RGB video from a unified latent space, we provide an additional interaction-oriented supervisory signal, guiding the model to generate physics-realistic interactions. 
%To improve the physical realism of the generated video, we leverage our HOI representation as auxiliary supervision within a novel joint-generation paradigm (see~\cref{fig:joint_generation}). Our approach trains the model to simultaneously synthesize both the RGB video ($V_\textnormal{RGB}$) and its corresponding HOI representation ($V_\textnormal{HOI}$) from a unified latent space. %This multi-task objective compels the model to learn the rich physical and structural priors from the HOI stream, which in turn regularizes the main video generation task and enhances its fidelity.
%Since our HOI representation is also video-based, this enables the diffusion model to synthesize both modalities simultaneously from a unified latent space. %This joint-generation process guarantees a cohesive generation of visual appearance and interaction dynamics.

\noindent \textbf{Unified Latent Space.} We adopt a 3D VAE to encode the RGB video ($V_{\textnormal{RGB}}$) and the HOI representation ($V_{\textnormal{HOI}}$) into a unified latent space, yielding visual tokens ($X_{\textnormal{RGB}}$) and interaction tokens ($X_{\textnormal{HOI}}$). The video-like structure of our HOI representation enables this joint encoding. Our model operates on this unified space by concatenating these tokens $X_\textnormal{RGB}$ and $X_\textnormal{HOI}$ into a single sequence $Z=(X_\textnormal{RGB}\oplus X_\textnormal{HOI})$. During training, $Z$ is corrupted to $Z_{t} = \sqrt{\bar\alpha_t}\, Z + \sqrt{1 - \bar\alpha_t}\,\varepsilon$, and our denoiser is trained to predict the added noise $\hat{\varepsilon}$ from $Z_t$. During inference, the denoiser reverses this process from pure noise $Z_T$ to recover the clean tokens $Z_0$, which are then decoded by the VAE to jointly generate the RGB video $\hat{V}_\textnormal{RGB}$ and its corresponding HOI representation $\hat{V}_\textnormal{HOI}$.

\noindent \textbf{Hierarchical Joint Denoiser.}
We introduce a hierarchical joint denoiser, built upon a Diffusion Transformer (DiT) backbone, to co-denoise visual
and interaction tokens, conditioned on a task description and an observed image (as shown in~\cref{fig:joint_generation}). The core of this denoiser is a novel share-and-specialization strategy. This strategy is motivated by the dual nature of these two modalities: they are inherently semantically coupled, yet each exhibits distinct modality-specific characteristics. We implement this strategy by partitioning the DiT layers into two cascaded modules.

Specifically, the \textit{Shared Semantic} module (layers 1 to ${k^*}$) enforces cross-modal consistency via our alignment loss, $L_\textnormal{align}$. This loss compels the hidden states of both the visual and interaction streams to align by explicitly maximizing the cosine similarity between their hidden states from layer ${k^*}$: %by explicitly maximizing the cosine similarity between the visual and interaction modalities via an alignment loss  on the hidden states from layer :
\begin{equation}
    L_\textnormal{align} =  \sum_{m=1}^{S} \left( 1 - \frac{H_{k^*}^m \cdot H_{k^*}^{S+m}}{\|H_{k^*}^m\| \|H_{k^*}^{S+m}\|} \right),
\end{equation}
where $H_{k^*}^m$ is the hidden state for the $m$-th visual token and $H_{k^*}^{S+m}$ is the hidden state for the corresponding $m$-th interaction token. Here, $S$ is the total number of visual tokens. This forced alignment compels the module to learn the shared interaction semantics between video and HOI representation. %This forced alignment acts as a strong cross-modal regularizer, compelling the video stream to learn the interaction semantics and physical priors captured by the HOI representation.
%The \textit{Specialized Details} module begins at layer ${k^*}+1$. Here, we introduce a learnable interaction embedding, $d_\textnormal{HOI}$.
Starting from layer ${k^*}+1$, the \textit{Specialized Details} module introduces a learnable interaction embedding, $d_\textnormal{HOI}$.
This embedding is added exclusively to the interaction token hidden states output by 
layer ${k^*}$ (i.e., $\tilde{H}_{k^*}^m = H_{k^*}^m + d_\textnormal{HOI}$ for $m > S$), 
before the subsequent DiT layers process them. This injection introduces a 
modality-specific bias that encourages the network to capture 
the unique characteristics required for each distinct stream. 

In addition, across both modules, each DiT layer adopts two architectural modifications for better adaptation to interaction tokens. For one thing, we assign identical positional encodings to the visual and interaction tokens sharing the same spatio-temporal position, thereby explicitly encoding their correspondence. Before the self-attention operation in layer $k$, the hidden states from the previous layer $H_{k-1}$ are modulated by these encodings, yielding the modulated tokens $P_k^m$:
\begin{equation}
    P^m_k = 
    \begin{cases}
        H_{k-1}^m + p^m, & \text{if } m \le S,
        \\
        %\quad \text{(Visual Token)} \\
        H_{k-1}^m + p^{m-S}, & \text{if } m > S. %\quad \text{(Interaction Token)}
    \end{cases}
\end{equation}
% The formula ensures that the interaction tokens ($m > S$) receive the same positional encoding ($p^{m-S}$) as their corresponding visual tokens ($m-S$).
For another, we integrate lightweight LoRA modules~\cite{hu2022lora} into the self-attention projection matrices ($\mathbf{W}_\mathcal{Q}, \mathbf{W}_\mathcal{K}, \mathbf{W}_\mathcal{V}$) for efficiency. To adapt the model for HOI generation while retaining its pre-trained visual knowledge, the LoRA update is applied selectively at the output stage using a binary mask. For each projection matrix $\mathbf{W}_{z}$ (where $z \in \{\mathcal{Q}, \mathcal{K}, \mathcal{V}\}$) and its corresponding low-rank adaptation $\text{LoRA}_{z}(\cdot)$, the projected output $\mathbf{X}_{z}^{\star}$ for the sequence ${P}_k$ is given by:
\begin{equation}
\mathbf{X}_{z}^{\star} = {P}_k \mathbf{W}_{z} + \gamma \cdot \operatorname{diag}(\mathbf{M}) \text{LoRA}_{z}(P_k).
\end{equation}
Here, $\gamma$ is a scaling factor and $\mathbf{M}\in\{0,1\}^{2S}$ is a binary mask (converted to its diagonal form) that activates LoRA updates only for interaction tokens. %This ensures the low-rank adaptation term \(\text{LoRA}_{z}(P_k)\) to only those positions corresponding to interaction tokens (i.e., where \(M_i = 1\)), while leaving all other positions unaffected. %$\mathbf{M} \in \{0, 1\}^{2S}$ is a binary mask vector where elements corresponding to interaction tokens are 1, and all others are 0. %This mask ensures the LoRA residual is exclusively added to the interaction token projections, enabling the model to learn the nuances of our HOI representation while fully leveraging the pre-trained weights for visual appearance.

\noindent \textbf{Training Loss.}
Our composite loss combines the alignment loss ($L_\textnormal{align}$) alongside diffusion losses for visual tokens ($L_\textnormal{RGB}$) and interaction tokens ($L_\textnormal{HOI}$):
\begin{equation}
    L = L_\textnormal{RGB} + \lambda_\textnormal{HOI} L_\textnormal{HOI} + \lambda_\textnormal{align} L_\textnormal{align}.
\end{equation}
Here, $L_\textnormal{RGB}$ and $L_\textnormal{HOI}$ can be formulated based on different prediction targets, such as the original noise~\cite{ho2020denoising} or velocity~\cite{liu2022flow}. The terms $\lambda_\textnormal{HOI}$ and $\lambda_\textnormal{align}$ balance the diffusion loss for interaction tokens and the alignment loss.

%% file: sec/4_experiment.tex
\section{Experiments}
\begin{table*}[!t]
  \centering
  \small
  \caption{Quantitative comparison with state-of-the-art methods on the Taste-Rob~\cite{zhao2025taste} and Taco~\cite{liu2024taco} datasets. All metrics are from VBench~\cite{huang2024vbench} and higher is better ($\uparrow$). Our method consistently outperforms prior work across all categories. The best results are in \textbf{bold}, while the second best are \underline{underlined}.}
  \label{tab:quantitative_comparison}
  \setlength{\tabcolsep}{4.9pt} % Adjust column spacing for readability
  \renewcommand{\arraystretch}{1.05}
  % The structure is now: 1 method column, then 6 metrics for each of the 2 datasets.
  \begin{tabular}{l ccc cc c ccc cc c}
    \toprule
    \multirow{3}{*}[-1.5ex]{Method}
      & \multicolumn{6}{c}{Taste-Rob~\cite{zhao2025taste}}
      & \multicolumn{6}{c}{Taco~\cite{liu2024taco}} \\
      \cmidrule(lr){2-7} \cmidrule(lr){8-13}
      & \multicolumn{2}{c}{Video Quality} & \multicolumn{2}{c}{Image Align.} & \multicolumn{1}{c}{Text. Align.} & \multicolumn{1}{c}{Overall}
      & \multicolumn{2}{c}{Video Quality} & \multicolumn{2}{c}{Image Align.} & \multicolumn{1}{c}{Text. Align.} & \multicolumn{1}{c}{Overall} \\
    \cmidrule(lr){2-3} \cmidrule(lr){4-5} \cmidrule(lr){6-6} \cmidrule(lr){7-7} \cmidrule(lr){8-9} \cmidrule(lr){10-11} \cmidrule(lr){12-12} \cmidrule(lr){13-13}
      & SC$\uparrow$ & IQ$\uparrow$ & ISC$\uparrow$ & IBC$\uparrow$ & VCS$\uparrow$ & TS$\uparrow$
      & SC$\uparrow$ & IQ$\uparrow$ & ISC$\uparrow$ & IBC$\uparrow$ & VCS$\uparrow$ & TS$\uparrow$ \\
    \midrule
    \textsf{CogVideoX}~\cite{yang2024cogvideox}
      & 0.959 & 0.688 & 0.955 & 0.954 & 0.187 & 8.959
      & 0.895 & 0.665 & 0.933 & 0.942 & 0.182 & 8.511 \\
    \textsf{Wan2.1}~\cite{wan2025wan}
      & 0.943 & \underline{0.700} & 0.947 & 0.939 & 0.185 &  8.897
      & 0.905 & \underline{0.717} & 0.933 & 0.947 & 0.189 & 8.792  \\
    \textsf{FLOVD}~\cite{jin2025flovd}
      & 0.941& 0.691 & 0.949 & 0.956& 0.189 & 8.888 
      &  0.903 & 0.686  & 0.927   & 0.947    & 0.177    & 8.619    \\
    \midrule
    
    $\textsf{SCAR}_{\textsf{C}}~\textsf{(our)}$
      & \textbf{0.964} & 0.696 & \underline{0.960} &\textbf{0.959} & \underline{0.193} & \underline{9.043} 
      & \textbf{0.916}& 0.698 & \textbf{0.951} & \textbf{ 0.954} & \underline{0.187} &\underline{ 8.793}  \\
     
    $\textsf{SCAR}_{\textsf{W}}~\textsf{(our)}$
      & \underline{0.961} & \textbf{0.709} &  \textbf{0.961}  & \underline{0.958} & \textbf{0.194} & \textbf{9.084 }
      & \underline{0.912} & \textbf{0.728} &  \underline{0.948} &  \underline{0.952} & \textbf{0.191} &  \textbf{8.899}  \\
    \bottomrule
  \end{tabular}
\end{table*}

\subsection{Experimental Setup}
\noindent \textbf{Datasets.} We conduct experiments on Taste-Rob~\cite{zhao2025taste} and Taco~\cite{liu2024taco} datasets. Taste-Rob is a large-scale fixed-view HOI dataset consisting of 100,856 videos, each paired with a manually annotated task description. This dataset features both single-hand and double-hand interactions, covering a diverse range of action categories (e.g., grasping, lifting, rotating), object types, and scene layouts. We follow Taste-Rob~\cite{zhao2025taste} to split training and test sets. 

We also conduct experiments on the Taco dataset. This dataset comprises 2,317 videos of double-hand interactions from both egocentric and third-person viewpoints. For our experiments, we only use the videos captured from the egocentric perspective, which inherently involves significant camera motion and complex background changes. We split the data into training (90\%) and testing (10\%) sets according to the dataset's triplet (action, tool, target object) categorization, ensuring each triplet is randomly split.

\noindent \textbf{Evaluation Criteria.}
We employ a set of video generation metrics from VBench~\cite{huang2024vbench} to assess three key aspects: (1) \textbf{Video Quality}, evaluated using \textit{subject consistency} (SC) and \textit{imaging quality} (IQ), (2) \textbf{Image-to-Video Alignment}, evaluated using \textit{i2v subject consistency} (ISC) and \textit{i2v background consistency} (IBC) to assess consistency with the observed image, and (3) \textbf{Text-to-Video Alignment}, evaluated using the \textit{viclip score} (VCS) for consistency with the task description. Finally, we report the weighted \textit{total score} (TS), which integrates these metrics based on the coefficients defined in VBench~\cite{huang2024vbench}.

\noindent \textbf{Implementation Details.}
To demonstrate the versatility of our method, we adapt our approach to two mainstream pre-trained video diffusion models (VDMs): CogVideoXI2V-5B and Wan2.1I2V-14B. We denote these two instantiations as $\textsf{SCAR}_\textsf{C}$ and $\textsf{SCAR}_\textsf{W}$, respectively. We preserve the core architecture of the base VDMs, including their original VAE latent space, number of DiT layers, and hidden state dimensions. 
For video generation length, we set the sequence length to 17 frames for the Taste-Rob~\cite{zhao2025taste} and 25 frames for the Taco~\cite{liu2024taco}. For $\textsf{SCAR}_\textsf{C}$, we set the LoRA dimension to 128 and apply an alignment loss to the hidden states of the 12th DiT layer. For $\textsf{SCAR}_\textsf{W}$, the LoRA dimension is increased to 256, and the alignment loss is applied to the 12th DiT layer. For both instantiations, the weights for the HOI diffusion loss, $\lambda_{\text{HOI}}$, and the alignment loss, $\lambda_{\text{align}}$, are consistently set to 1.0 and 0.1, respectively.

\begin{figure*}[!t]
  \centering
  \includegraphics[width=0.98\textwidth]{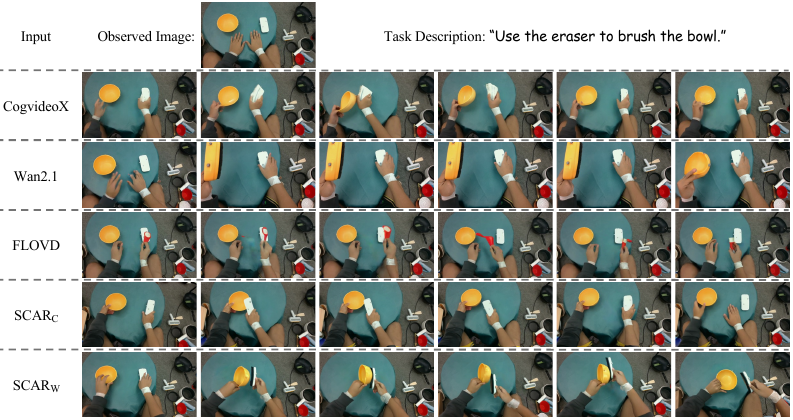} % Replace with your image file path
  \caption{Qualitative comparison with state-of-the-art methods on Taco~\cite{liu2024taco} dataset.\textsf{CogVideoX}~\cite{yang2024cogvideox} produces distorted hands and implausible contact. \textsf{Wan2.1}~\cite{wan2025wan} fails to generate the semantically correct action described in the task description. The two-stage \textsf{FLOVD}~\cite{jin2025flovd} suffers from error propagation, where inaccurate initial optical flow results in hallucination (a red object suddenly appearing). In contrast, our \textsf{SCAR} generates physics-realistic, temporally coherent videos by jointly generating our proposed HOI representation. Please refer to the supplementary video for better illustration. Experimental results on the Taste-Rob~\cite{zhao2025taste} are also available in the supplementary material.}
  \label{fig:qualitative_comparison}
\end{figure*}

\subsection{Comparisons with State-of-the-art Methods}
\begin{figure}[!t]
  \centering
  \includegraphics[width=0.475\textwidth]{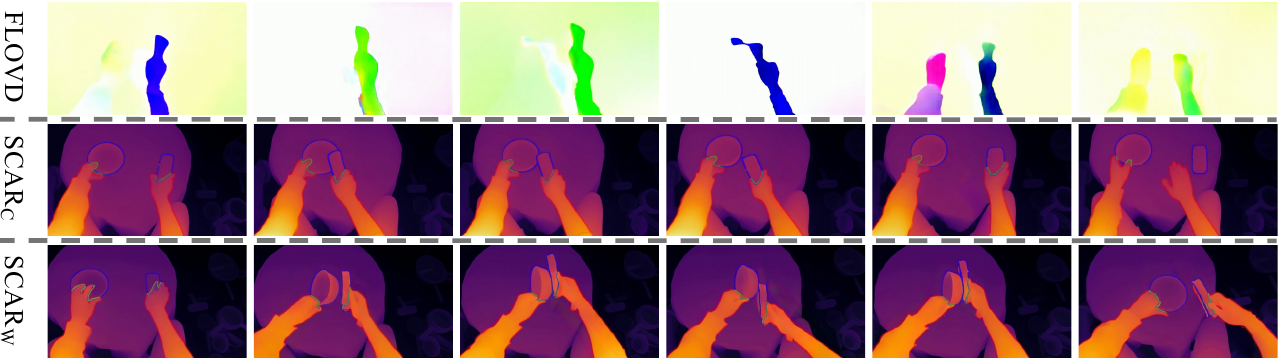} % Replace with your image file path
  %\caption{Visualization of the HOI representations corresponding to the results shown in~\cref{fig:qualitative_comparison}. FLOVD~\cite{jin2025flovd} relies on optical flow, which can be noisy and inaccurate, leading to the error propagation seen in the final video. In contrast, our proposed HOI representation consists of depth maps and contact-augmented contours. This representation provides stable scene structure (e.g., consistent depth for the bowl) and precise interaction cues (hand contours clearly brushing the bowl), which enables the generation of coherent and realistic videos.}
\caption{Qualitative comparison of generated HOI representations corresponding to~\cref{fig:qualitative_comparison}. The optical flow generated by \textsf{FLOVD}~\cite{jin2025flovd} is noisy and inaccurate, which leads to the error propagation seen in the final video. In contrast, our jointly generated representation embodies consistent structural and contact cues, indicating that the model captures physical interaction patterns.} %FLOVD~\cite{jin2025flovd} relies on a separate optical flow stage, where initial inaccuracies and noise propagate to the final video. In contrast, our representation is jointly generated with the video in an end-to-end manner. This joint generation process itself acts as a structure and contact-aware supervisory signal. By compelling the model to generate this HOI representation, it simultaneously learns to produce a coherent and physics-realistic video.}
  \label{fig:qualitative_comparison_hoi}
\end{figure}

\noindent \textbf{Methods for Comparison.} We compare our two instantiations, $\textsf{SCAR}_\textsf{C}$ and $\textsf{SCAR}_\textsf{W}$, against state-of-the-art video generation models. These baselines comprise two categories. First, we conduct a direct evaluation against their own underlying backbones, the general-purpose models \textsf{CogVideoX}~\cite{yang2024cogvideox} and \textsf{Wan2.1}~\cite{wan2025wan}. Second, we compare against \textsf{FLOVD}~\cite{jin2025flovd}, a representative two-stage method. To maintain fairness (especially against our $\textsf{SCAR}_\textsf{C}$), we specifically fine-tune its CogVideoX-based instead of SVD-based instantiation. To ensure a fair comparison, all methods (both our instantiations and the baselines) are fine-tuned on the same dataset splits and initialized from the same pre-trained model checkpoints. Further implementation details for all methods are provided in the supplementary material. %We conduct a comprehensive comparison of our method against two categories of video generation baselines: state-of-the-art general-purpose models (\textsf{CogVideoX}~\cite{yang2024cogvideox} and \textsf{Wan2.1}~\cite{wan2025wan}) and a representative two-stage method \textsf{FLOVD}~\cite{jin2025flovd}, which first generates optical flow as an intermediate control signal. For a fair comparison, all baselines are fine-tuned under the same dataset splittings. We provide additional implementation details for these baselines in the supplementary material. 

\noindent \textbf{Results.} %As illustrated in~\cref{tab:quantitative_comparison}, our SCAR outperforms all state-of-the-art methods across all quantitative metrics on both Taste-Rob~\cite{zhao2025taste} and Taco~\cite{liu2024taco} datasets. We attribute this success to our proposed structure and contact-aware representation and joint-generation paradigm, which resolve key failure modes present in prior work.
As depicted in~\cref{fig:qualitative_comparison}, general I2V models without HOI representation, \textsf{CogVideoX}, often struggle with physics realism, generating videos with distorted hands and objects, and unrealistic contact. This observation is consistent with their lower video quality scores (SC and IQ) reported in~\cref{tab:quantitative_comparison}. Although \textsf{Wan2.1} produces higher video quality, its failure to align with the task description (e.g., failing to brush the bowl as the task description) directly results in a low text-to-video alignment score (VCS). The two-stage model \textsf{FLOVD} suffers from error propagation, where inaccuracies from the initial optical flow (visualized in~\cref{fig:qualitative_comparison_hoi}) are compounded in the second stage, resulting in temporal inconsistencies and hallucinated objects (e.g., the red object). This failure to maintain object identity is quantitatively reflected in its poor performance on the image-to-video alignment metrics (particularly ISC). 
In contrast, our \textsf{SCAR} framework generates physics-realistic and temporally coherent videos that execute the ``brushing'' action (see last two rows in~\cref{fig:qualitative_comparison}), achieving superior performance across all metrics in~\cref{tab:quantitative_comparison}. The high fidelity of the generated HOI representation (see~\cref{fig:qualitative_comparison_hoi}) demonstrates that the model has captured the underlying physical constraints and interaction patterns for physics-realistic video synthesis. %This success stems from the explicit structure and contact constraints provided by our jointly-generated HOI representation (see~\cref{fig:qualitative_comparison_hoi}). 
Notably, consistent performance gains of both $\textsf{SCAR}_\textsf{C}$ and $\textsf{SCAR}_\textsf{W}$ over baselines further demonstrate the versatility of our approach. Additionally, we provide more qualitative comparisons (including results on the Taste-Rob~\cite{zhao2025taste} dataset) in the supplementary material. 
%In contrast, our SCAR framework overcomes these limitations through the joint-generation paradigm and proposed structure and contact-aware HOI representation. By jointly synthesizing this representation, the generation model captures the essence of the interaction, where contact-augmented hand-object contours model the hand-object motion, and hand-object contact and depth maps provide a holistic structure context for the scene. This synergy mitigates the shape distortions and implausible contact seen in baselines, enabling the generation of physically and semantically correct interactions. As shown in~\cref{fig:qualitative_comparison}, our SCAR successfully generates a physics-realistic and temporally coherent video that executes the "brushing" action. It is also quantitatively validated by our superior scores across all metrics in~\cref{tab:quantitative_comparison}. Notably, consistent performance gains of both $\text{SCAR}_\text{C}$ and $\text{SCAR}_\text{W}$ over baselines further demonstrate the robustness and versatility of our approach.

\subsection{Generalization to Open-world Scenarios}
\label{sec:in_the_wild}
\begin{figure*}
    \centering
    \includegraphics[width=0.98\linewidth]{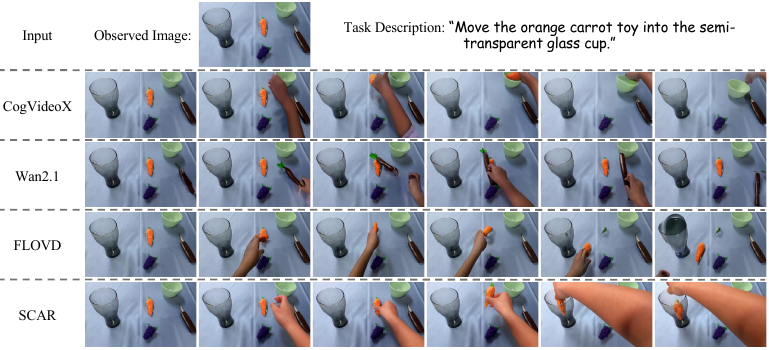}
    \caption{Qualitative comparison on a challenging open-world task. \textsf{CogVideoX}~\cite{yang2024cogvideox}, \textsf{Wan2.1}~\cite{wan2025wan}, and \textsf{FLOVD}~\cite{jin2025flovd} fail to follow the task description and produce physically implausible interactions. In contrast, our \textsf{SCAR} generates a physics-realistic, coherent video that correctly executes the challenging task involving unseen target objects and distractors.}
    \label{fig:in_the_wild_main}
\end{figure*}
To evaluate the open-world generalization of our method beyond the closed-set environments, we collect a new benchmark of 200 challenging open-world samples. Each sample consists of a task description and an image containing unseen target objects. Without loss of generality, we use our $\textsf{SCAR}_\textsf{W}$ trained on the Taste-Rob~\cite{zhao2025taste} dataset for this evaluation. 

As demonstrated in \cref{fig:in_the_wild_main}, we present a representative qualitative comparison in a challenging open-world scenario, where the entire scene 
consists of unseen objects (e.g., the semi-transparent cup, carrot, green bowl) that act as complex distractors for one another. All baseline methods (\textsf{CogVideoX}~\cite{yang2024cogvideox}, \textsf{Wan2.1}~\cite{wan2025wan}, \textsf{FLOVD}~\cite{jin2025flovd}) exhibit significant hand-object 
distortion, temporal inconsistencies, and physically 
implausible contact. Furthermore, their generated videos often fail to 
align with the task description. Specifically, 
\textsf{CogVideoX} places 
the carrot into the green bowl instead of the 
``glass cup''. \textsf{Wan2.1} exhibits poor object grounding, picking up the wrong item entirely. \textsf{FLOVD} 
fails to follow the precise instruction, moving the carrot ``towards''
rather than ``into'' the cup. These qualitative failures are corroborated by relatively low metrics (see the supplementary material). These baseline methods, even with large-scale finetuning, still struggle with generalization as they have difficulty capturing complex interaction physics.
By comparison, our \textsf{SCAR} generates physics-realistic and temporally coherent videos in challenging open-world scenarios, achieving superior performance across all metrics. This success highlights that our method captures interaction patterns consistent with physical constraint by learning to jointly predict the video and our structure and contact-aware representation. We provide more qualitative comparisons in the supplementary material.
\begin{table}[!t] 
  \centering
  \small
  \caption{Quantitative comparison of different HOI representations. For all metrics, higher is better ($\uparrow$). The best results are in \textbf{bold}.}
  \label{tab:ablation_comparison}
  \renewcommand{\arraystretch}{1.02} % 1.2的行高更适合booktabs

  \setlength{\tabcolsep}{0pt} % <-- 您要求的
  \begin{tabular*}{\columnwidth}{l @{\extracolsep{\fill}} ccccc}
    \toprule
    \multirow{2}{*}{HOI Rep.} & \multicolumn{2}{c}{Video Quality} & \multicolumn{2}{c}{Image Align.} & \multicolumn{1}{c}{Text Align.}\\
    \cmidrule(lr){2-3} \cmidrule(lr){4-5} \cmidrule(lr){6-6}
    & SC $\uparrow$ & IQ $\uparrow$ & ISC $\uparrow$ & IBC $\uparrow$ & VCS $\uparrow$ \\
    \midrule
    
    % --- 绿色组 (只为第一列上色) ---
    \cellcolor{PastelGreen}OF    & 0.889 & 0.660 & 0.935 & 0.942 & 0.177 \\
    \cellcolor{PastelGreen}HOM   & 0.903 & 0.689 & 0.939 & 0.945 & 0.181 \\
    \cellcolor{PastelGreen}DM   & 0.889 & 0.682 & 0.940 & 0.944 & 0.180 \\
    \midrule
    
    % --- 黄色组 (只为第一列上色) ---
    \cellcolor{PastelYellow}w/o HOC & 0.899 & 0.689 & 0.937 & 0.945 & 0.181 \\
    \cellcolor{PastelYellow}w/o CG  & 0.906 & 0.687 & 0.945 & 0.948 & 0.179 \\
    \cellcolor{PastelYellow}w/o DM  & 0.901 & 0.690 & 0.939 & 0.941 & 0.180 \\
    \cellcolor{PastelYellow}+ KP    & 0.891 & 0.691 & 0.940 & 0.943 & 0.183 \\
    \midrule
    
    \textbf{SCAR (our)} & \textbf{0.916} & \textbf{0.698} & \textbf{0.951} & \textbf{0.954} & \textbf{0.187} \\
    \bottomrule
  \end{tabular*}
\end{table}
\subsection{Analysis of HOI Representation Design}
\label{sec:ana_model}
In this subsection, we evaluate the effectiveness of our HOI representation design by comparing our full model against two categories of variants (see \cref{tab:ablation_comparison}). Without loss of generality, all experiments for this analysis are conducted on the Taco~\cite{liu2024taco} dataset, using \textsf{SCAR}$_\textsf{C}$ as our full model.

\noindent \textbf{Comparison with Existing Representation.}
The first category (highlighted in \colorbox{PastelGreen}{green}) substitutes our proposed representation with existing representations, including optical flow (OF), hand-object masks (HOM), and depth maps (DM). As shown in~\cref{tab:ablation_comparison}, these representations, which capture only one aspect of the interaction, lead to reduced performance. For instance, the OF and DM variants fail spatiotemporal coherence, causing objects (e.g., the blue ruler) to disappear, while the HOM variant lacks explicit contact cues, leading to missed grasps (see supplementary material for qualitative comparison).

\noindent \textbf{Ablation Study on Proposed Representation.}
The second category (highlighted in \colorbox{PastelYellow}{yellow}) consists of variants ablating our proposed components: w/o HOC (removing hand-object contours), w/o CG (removing contact region), w/o DM (removing depth maps), and +KP (adding rendered 2D hand keypoints). As shown in~\cref{tab:ablation_comparison}, removing any of our core components (w/o HOC, w/o CG, w/o DM) degrades performance, confirming their complementary importance. The qualitative comparison (see supplementary material) reveals that variants lacking spatial localization (w/o HOC) or holistic structure (w/o DM) struggle with object consistency, and the w/o CG variant fails fine-grained tasks (e.g., measuring cup). Furthermore, adding 2D keypoints (+KP) also degrades performance, as an overly complex auxiliary generative target hinders optimization.

In summary, compared with the above existing representations and our ablated variants, our SCAR achieves the best quantitative metrics shown in~\cref{tab:ablation_comparison} and generates more physically realistic videos (see the supplementary material). This demonstrates that our HOI representation benefits from a more comprehensive and interaction-oriented supervisory signal, which effectively guides the learning of fine-grained interaction physics.

%% file: sec/5_conclusion.tex
\section{Conclusion}
In this work, we introduce an open-world hand-object interaction video generation framework based on a structure and contact-aware representation. 
We introduce a novel HOI representation that encodes hand-object contact, the spatial localization of the hand and object, and the holistic structure of the observed scene. This representation is leveraged within a joint-generation paradigm featuring a share-and-specialization strategy that mitigates the error accumulation of prior multi-stage methods. We further curate the above representation for more than 100K HOI videos to enable large-scale training. 
Experiments demonstrate that our method substantially outperforms prior work in physical realism and temporal consistency, and strong generalization to challenging open-world scenarios. %Our framework provides a scalable path toward high-fidelity synthesis of contact-rich interactions for applications in robotics and simulation.